\documentclass[journal]{IEEEtran}
\usepackage{amsmath,amsfonts}
\usepackage{algorithmic}
\usepackage{algorithm}
\usepackage{array}
\usepackage[caption=false,font=normalsize,labelfont=sf,textfont=sf]{subfig}
\usepackage{textcomp}
\usepackage{stfloats}
\usepackage{url}
\usepackage{verbatim}
\usepackage{graphicx}
\usepackage{cite}
\usepackage{amssymb}
\newtheorem{definition}{Definition}

\begin{document}

\title{Robustness of Object Detection of Autonomous Vehicles\\ in Adverse Weather Conditions}

\author{Fox Pettersen and Hong Zhu,~\IEEEmembership{IEEE Senior Member}
\thanks{Fox Pettersen and Hong Zhu are with the School of Engineering, Computing and Mathematics, Oxford Brookes University, Oxford, UK.}
\thanks{Correspondence Email: hzhu@brookes.ac.uk.}
}

\markboth
{F. Pettersen and H. Zhu: Robustness of Object Detection in Adverse Weather Conditions}
{}

\maketitle

\begin{abstract}
As self-driving technology advances toward widespread adoption, determining safe operational thresholds across varying environmental conditions becomes critical for public safety. This paper proposes a method for evaluating the robustness of object detection ML models in autonomous vehicles under adverse weather conditions. It employs data augmentation operators to generate synthetic data that simulates different severance degrees of the adverse operation conditions at progressive intensity levels to find the lowest intensity of the adverse conditions at which the object detection model fails. The robustness of the object detection model is measured by the average first failure coefficients (AFFC) over the input images in the benchmark. The paper reports an experiment with four object detection models: YOLOv5s, YOLOv11s, Faster R-CNN, and Detectron2, utilising seven data augmentation operators that simulate weather conditions fog, rain, and snow, and lighting conditions of dark, bright, flaring, and shadow. The experiment data show that the method is feasible, effective, and efficient to evaluate and compare the robustness of object detection models in various adverse operation conditions. In particular, the Faster R-CNN model achieved the highest robustness with an overall average AFFC of 71.9\% over all seven adverse conditions, while YOLO variants showed the AFFC values of 43\%.  The method is also applied to assess the impact of model training that targets adverse operation conditions using synthetic data on model robustness. It is observed that such training can improve robustness in adverse conditions but may suffer from diminishing returns and forgetting phenomena (i.e., decline in robustness) if overtrained.
\end{abstract}


\section{Introduction}

\IEEEPARstart{R}{ecent} advancements in autonomous driving technology have accelerated the development of self-driving vehicles, with applications ranging from passenger transport to logistics and delivery. Despite this progress, one of the challenges facing the industry is determining under which environmental conditions these vehicles can operate safely. Adverse weather conditions, such as rain, fog, snow, or low-light scenarios, can significantly reduce the performance of object detection models, which are vital for the autonomous vehicle to navigate and avoid obstacles. Establishing clear safe operation thresholds is therefore essential to ensure that autonomous vehicles only operate when it is safe to do so.

However, there is currently no standardised way to measure the robustness of a machine learning (ML) model to define the safe operation boundary. A reliable evaluation methodology is needed to assess how well such ML models perform as weather worsens and to determine the point at which object detection begins to fail below the required safety level. Such a method would support the broader industry goal of identifying safe operational limits for autonomous object recognition models. It would also validate the effectiveness of weather-augmented training of object detection models. 

This paper proposes a new robustness testing and evaluation method and applies it to the robustness of object detection models used in autonomous vehicles under adverse operation conditions. It will identify the weather intensity thresholds at which object detection performance becomes unreliable, providing safety-critical insights into when autonomous systems should be restricted or disengaged. The paper also reports an experiment in assessing the effectiveness of training machine learning models with weather-augmented training data in improving model resilience, offering guidance for ML model developers and manufacturers seeking to improve operational safety.

The paper is organised as follows. Section II briefly reviews related work. Section III presents the proposed method and reports the implementation of the method as a testing tool. Section IV reports the experiments with the proposed method. Section V concludes the paper with a discussion of the directions for future work.

\section{Related Works}

\subsection{Object Detection in Autonomous Vehicles}

The use of autonomous vehicles relies on robust object detection models capable of identifying objects in real-time driving environments \cite{liu2020creating}. Object detection approaches utilise deep learning architectures, with convolutional neural networks (CNNs) forming the essential part of most models. The evolution from traditional computer vision techniques to deep learning-based methods has significantly improved detection accuracy; however, questions remain about their robustness under adverse weather conditions.

YOLO\footnote{URL: https://github.com/ultralytics \label{fnt:YOLOGitHUb}} architectures represent a significant advancement in real-time object detection, prioritising speed alongside accuracy through single-stage detection \cite{redmon2016you}. The progression from YOLOv5 to YOLOv11 demonstrates iterative improvements in both architecture and training methodologies \cite{sapkota2025ultralytics}. In contrast, two-stage detectors such as Faster R-CNN prioritise accuracy through proposal networks, achieving higher precision at the cost of speed \cite{ren2015faster}.

\subsection{Weather-Related Performance Degradation}

It is widely recognised that adverse weather conditions present substantial challenges to object detection systems, with fog, rain, snow, and lighting interference degrading image quality \cite{bijelic2020seeing}. Existing research has reported significant performance reductions under adverse environmental conditions.

For example, U{\v{r}}i{\v{c}}{\'a}{\v{r}} et al. reported that rain affects object detection through water droplets on camera lenses and road surface reflections \cite{uvrivcavr2019soilingnet}. Sakaridis, Dai and Van Gool studied how fog presents distinct challenges through uniform light reduction and loss of depth perception \cite{sakaridis2018semantic}. Bijelic et al. demonstrated that a deep multimodal sensor can maintain performance in unseen adverse weather conditions \cite{bijelic2020seeing}. Lighting extremes, including darkness, excessive brightness, and sun flare, challenge the dynamic range limitations of the physical camera sensors. Dai and Van Gool explored dark model adaptation for image segmentation from daytime to nighttime conditions, highlighting the significant performance gaps that exist across lighting conditions \cite{dai2018dark}. 

\subsection{Performance and Robustness of Object Detection}

\begin{figure}[!h]
\centering
\includegraphics[width=3.4in]{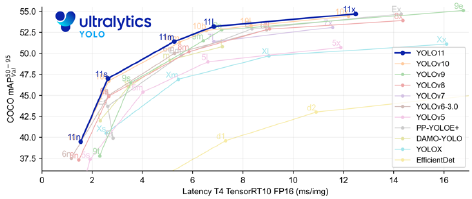}
\caption{Performances of YOLO Models Measured by mAP$^{\ref{fnt:YOLOGitHUb}}$}
\label{fig:YOLOPerformance}
\end{figure}

Performance metrics, including mean Average Precision (mAP), Area under the Precision-Recall Curves (AoC), and Intersection over Union (IoU), have been widely used in the evaluation of the performance of ML models for object detection, which measure the correctness and reliability of object detection. For example, Fig. \ref{fig:YOLOPerformance} shows the performances of various YOLO object detection models measured by mAP$^{\ref{fnt:YOLOGitHUb}}$. However, such performance metrics fail to capture the robustness of ML models, i.e. how well ML models perform under adverse operational conditions. 

Several researchers have proposed corruption benchmarks to evaluate model robustness. In 2019, Hendrycks and Dietterich proposed a framework for benchmarking neural network robustness to common image corruptions \cite{hendrycks2019benchmarking}. Also in 2019, Michaelis et al. specifically addressed robustness in object detection for autonomous driving environments, creating a benchmark by applying various corruption types to test datasets for evaluation of autonomous driving in winter weather condition \cite{michaelis2019benchmarking}. These benchmarks typically apply standardised corruptions to test datasets with fixed intensity levels and corruption types. Such benchmarks can be used to evaluate the likelihood that an ML model fails in the specific intensity level of corruptions as represented by the benchmark. However, such benchmarks cannot determine to what intensity level of corruptions the model could still perform as well as expected. An approach to identifying the specific threshold of corruption intensity below which an ML model is safe to operate remains absent. 

In a different context, adversarial robustness of neural networks has been an active research area in the past decade since Szegedy et al.'s discovery of the universal existence of adversarial examples in deep neural networks (DNN) for object recognition and identification in 2014 \cite{szegedy2013intriguing}. An adversarial example of a natural image $\neg x$ for a DNN M for image recognition, an image $x'$ such that $M(x)\neq M(x')$ but $x'$ is not perceptually distinguishable from $x$  by human beings. A large number of algorithms for generating adversarial examples have been proposed with different adversarial attack models, such as white-box and black-box attacks; see \cite{khamaiseh2022adversarial} for a recent survey by Khamaiseh et al. in 2022. 

Given an adversarial example generation algorithm $G(x,M)$, which for any given ML model $M$ produces an adversarial example $x'=G(x,M)$  from an input image $x$, a metric for measuring the adversarial robustness of an ML model $M$ is
\[R_G (T,M)=\frac{1}{|T|}  \sum_{x \in T}\|x,G(x,M)\| \]
where $T$ is the test dataset of the benchmark used for evaluation of the ML model $M$, $\|x,x'\|$ is a distance metric on the input data space of the ML model. It approximates the following theoretical definition of adversarial robustness, as given by Moosavi-Dezfooli et al. in 2016 \cite{moosavi2016deepfool}.  
\[\rho(M)=E_x \Delta(x,M),\]
where $E_x$ denotes the expectation over the data distribution, $\Delta(x,M)$ is the robustness of $M$ at the point $x$, which is defined as follows. 
\[ \Delta(x,m) = min \{ \|x,x'\| ~|~ M(x) \neq M(x') \}. \]

Informally, $\Delta(x,M)$ defines the safe zone surrounding the input point $x$ in the sense that the model does not fail on any data point whose distance to x is less than $\Delta(x,M)$. In other words, $\Delta(x,M)$ is the threshold for safe operation under adversarial attacks. The adversarial robustness is defined by  $\rho(M)$ and measured by the metric $R_G (T,M)$ on a benchmark test dataset $T$ is the expected size of the safe zone. 

However, adversarial robustness does not answer the research question about how to determine the threshold for the safe operation of object detection models in adverse weather and lighting conditions. This is because adversarial examples typically do not represent input under such conditions. Nevertheless, the notion of operational robustness under adverse conditions and the metric proposed in this paper are inspired by this. 

\subsection{Weather Augmentation in Training Data}

Data augmentation that artificially generates weather effects represents a common approach to improving model performance in adverse weather conditions. Various software tools and algorithms exist for simulating weather conditions in images, ranging from simple image processing operations to sophisticated neural rendering techniques. For example, Hahner et al. demonstrated semantic understanding of foggy scenes using purely synthetic data, suggesting that artificially generated weather conditions can effectively train models for real-world scenarios \cite{hahner2019semantic}. 
However, the effectiveness of weather-augmented training remains debatable. Some studies report significant improvements in performance in adverse weather conditions by including weather-interference images during training. For example, Kenk and Hassaballah developed the DAWN (Detection in Adverse Weather Nature) dataset, which showed positive results for vehicle detection in adverse weather conditions when models were trained with weather-augmented data \cite{kenk2020dawn}. Zhu et al. used seven data augmentations to generate synthetic images of various weather and lighting effects in training the object detection models of autonomous racing cars to improve the performance in adverse operating conditions \cite{zhu2023scenario}. 

However, other research suggests that adding synthetic weather can provide limited benefits or even degrade performance on clean data. The transfer between synthetically generated and real-world weather effects represents a persistent challenge, with questions about whether digitally generated rain, fog, or snow adequately replicates the physical processes affecting real sensor inputs.

\subsection{Safety Standards and Operational Limits}

The autonomous vehicle industry currently lacks standardised safety thresholds for object detection performance under adverse weather. While various governments have proposed testing requirements, few specify quantitative performance benchmarks across different environmental conditions. In 2021, the On-Road Automated Driving (ORAD) Committee has established definitions for terms related to driving automation systems, providing a framework for categorising autonomous capabilities, but specific weather-related operational limits remain undefined \cite{on2021taxonomy}. This regulatory gap creates challenges for manufacturers seeking to validate system safety and for insurance companies assessing liability.

Human driver capabilities under adverse weather provide an implicit baseline for autonomous system requirements. Sivak demonstrated that drivers rely on approximately 90\% visual information when operating vehicles, establishing the critical importance of object recognition modules \cite{sivak1996information}. However, calculating human performance metrics for machine perception proves a difficult challenge, as humans and object recognition modules exhibit different failure modes and capabilities.

\subsection{Gaps in Existing Research}

Despite substantial research in object detection models in autonomous vehicles, several critical gaps remain. 
\begin{itemize}
\item Synthetic data generated via data augmentations is widely used in model training and the construction of benchmarks for testing and evaluating object detection models in adverse operating conditions. However, such tests and evaluations are limited to using benchmarks that contain test data of a small number of fixed intensity levels rather than continuous intensity progressions, limiting understanding of gradual performance degradation. 

\item Comparisons across multiple model architectures under identical testing conditions remain mostly on performance, making it difficult to identify advantages in robustness. 

\item Existing metrics focus primarily on performance or adversarial robustness rather than identifying the specific boundaries for safe operations of autonomous vehicles under adverse weather and lighting conditions.  We lack a method to identify the point where a model transitions from reliable to unreliable operations under adverse conditions. 
 
\end{itemize}

This paper addresses these gaps by proposing a testing and evaluation methodology that measures the robustness of object detection across a continuous range of weather intensity, enabling identification of safe operation boundaries in adverse weather and lighting conditions.  The method can also be applied to the comparison of multiple models via testing them under identical situations and the evaluation of the effectiveness of weather-specific training by comparing the models before and after such training. Experiments will be reported to provide empirical evidence of the effectiveness of the proposed method.

\section{The Proposed Method}

This section presents the proposed method. We will first introduce the notion of operational robustness under a type of adverse conditions. Then, we define a metric to approximate the notion through experiments. Finally, we will describe an algorithm to measure the operational robustness using the metric in an experiment.

\subsection{The Notion of Operational Robustness}

Let $x$ be any given input image that represents a normal operation condition $x$. We assume that $\Phi(x)$ is the set of images surrounding $x$ that represent the situations in the type of adverse condition, but deviated from the normal operation condition x, such as in fog. The images $x'$ in $\Phi(x)$ may have different levels of severity or intensity of the adverse condition. We will write $\|x,x'\| \in [0,1]$ to denote the severity/intensity of the adverse condition, where, when the severity/intensity equals 0, the image is identical to the original, thus not under the adverse condition, while when the severity/intensity equals 1, it is at the maximal possible level of severity. 

Let $M$ be an object detection ML model such that $M(x)=\left< c,p\right>$, where $p$ is an area in $x$, which is typically represented by $\left<(a,b),w,h\right>$ for the coordinates $(a,b)$ of the upper-left corner of the area and its width $w$ and height $h$,  $c$ is the class of object at the area $p$. 

Let $x$ and $x'$ be two images. We write $M(x) \approx_\delta M(x')$ to denote that the ML model $M$ produces the same object detection results, which is formally defined as follows. Let $M(x)=\left<c,p\right>$ and $M(x')=\left<c',p'\right>$.
\[ M(x) \approx_\delta M(x') \Leftrightarrow c=c' \wedge p \cap p'> \delta \]
where $p \cap p'$ denotes the overlaps between two areas $p$ and $p'$, $0<\delta<1$  is a given real number denoting the tolerable error of the detection area. For example, $p \cap p'=0.99$ means the overlaps between $p$ and $p'$ is 99\%. For the sake of simplicity, we will ignore the subscript $\delta$ in the remainder of the paper without loss of generality. We will also write $M(x) \nsim M(x')$ for $\neg (M(x) \approx_\delta M(x'))$ to denote that the ML model $M$ produces different object detection results on inputs $x$ and $x'$. 

Now, we define the notion of operational robustness under a type of adverse conditions. 

\begin{definition}(Operational Robustness under A Type of Adverse Conditions) 

The \emph{operational robustness} of the ML model $M$ under a type $\Phi$ of adverse conditions, denoted by $\rho_\Phi(M)$, is defined as follows. 
\[\rho_\Phi(M)=E_x \Delta_\Phi(x,M)\]
where $E_x$ is the expectation over the data distribution,  $\Delta_\Phi(x,M)$ is the operational robustness under weather conditions $\Phi$ at point $x$, which is formally defined by the following equation. 
\[ \Delta_\Phi(x,M)= Min \{ \|x,x'\|~|~ x \in \Phi(x) \wedge M(x) \nsim M(x')\}.\]
\end{definition}

Informally, $\Delta_\Phi(x,M)$ is the minimal severance/intensity of the adverse condition that is below the intensity the ML model $M$ will not fail even if the input is in the adverse condition of type $\Phi$. Similar to $\Delta(x,M)$ in the definition of the notion of adversarial robustness, $\Delta_\Phi(x,M)$ defines a safe zone surrounding a normal point $x$. $\rho_\Phi(M)$ is therefore the expected size of the safe zone of the adverse condition in which the ML model will fail.  

\subsection{Metric for Measuring Operational Robustness}

It is difficult to measure operational robustness under a type of adverse conditions directly because such a benchmark is difficult to construct. However, with data augmentation techniques that generate synthetic data that simulate adverse operation conditions, measuring operation robustness is feasible. 

Assume that $\phi(x,i)$ be a data augmentation operator that generates a synthetic image $x'=\phi(x,i)$ that represents an adverse condition in $\Phi(x)$, where the parameter $x$ is the input image $x$ that represent a normal operation condition, the parameter $i\in [0,1]$ is called the interference strength, which represents the severity or intensity of the adverse condition that the effect of the adverse condition is injected into $x$. In other words, for all images $x$ of the normal operation condition, we have the following conditions. 
\[ \phi(x,i)\in \Phi(x) \wedge \|x,\phi(x,i)\|=i.\]

Let $T$ be the test dataset of the benchmark that each $x \in T$ is an image in the normal operation condition. The operational robustness is defined in Definition 1 can be measured by the metric $AFFC_\phi (T,M)$  defined as follows. 
\[AFFC_\phi(T,M)=\frac{1}{\|T\|}  \sum_{x \in T}FFC_\phi(x,M) \]
where  $FFC_\phi (x,M)$ is an ML model $M$’s first failure coefficient at point $x$, which is formally defined by the following equation. 
\[ FFC_\phi (x,M)=Min\{i_x  |M(x) \nsim M(\phi (x,i_x))\}.\]

\subsection{Search Algorithms}

The following is the linear search algorithm to find the first failure coefficients over a test dataset and calculate their average to measure the operational robustness. 

\begin{figure}[!h]
\label{fig:SearchAlgorithm}
\includegraphics[scale=1.2]{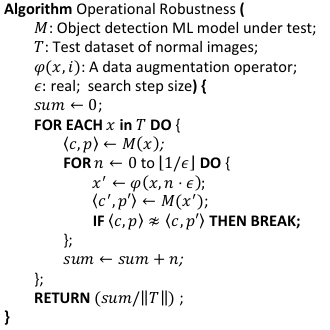}
\end{figure}

Note that, if the ML model under test fails monotonically with respect to the severance of adverse conditions, a more efficient algorithm, such as binary search, can be used. The monotonic condition can be formally stated as follows. 

\begin{definition} (Monotonicity of Failures)

An object detection model M is monotonic in failures with respect to the data augmentation operator $\phi$, if for all images $x$ of normal operation conditions, we have that for all $0<i<1$, 
\[ M(x) \nsim M(\phi(x,i)) \Rightarrow \forall j>i.(M(x) \nsim M(\phi(x,j)),\]
and
\[M(x) \approx M(\phi(x,i)) \Rightarrow \forall j<i.(M(x) \approx M(\phi(x,j)).\]
\end{definition}

However, it is unclear if object detection ML models are monotonic in failures. 

\section{Experiments}

Experiments have been conducted with the proposed method to answer the following research questions. 

\begin{description}
\item[RQ1:] Is the proposed method effective in evaluating and comparing the operational robustness of object detection models in various adverse weather conditions? 
\item[RQ2:] Is it effective to improve ML models’ operational robustness in adverse weather conditions by training ML models with synthetic data? 
\end{description}

To address these two research questions, two experiments were designed and conducted. The following two subsections report these experiments, respectively. 

\subsection{Evaluation And Comparison of Operational Robustness}

To answer research question \emph{RQ1}, we apply the proposed method to a number of ML models of object detection of the current state of the art and compare their robustness. 

\subsubsection{Design of the Experiment}

A sample of 4 pretrained object detection models of different varieties was selected as the subject ML models, which are: YOLOv5s, YOLOv11s, Faster R-CNN, and Detectron2. These models were selected because they are publicly available, pre-trained on automotive datasets, and represent different architectural approaches, the representation of the current state of art in ML object detection and are suitable for autonomous vehicles. These models were downloaded from GitHub. 

Due to the limitation of our resources, a total of 100 images will be used as our benchmark, encompassing a range of driving scenarios that a model would encounter. Among these images, 60 were sourced from dashcam footage, which gives us an accurate perspective of a forward-facing camera used in self-driving vehicles. The other 40 of them were sourced from the Common Objects in Context (COCO) dataset \cite{lin2014microsoft}. 

\begin{figure*}[ht]
\centering
\includegraphics[scale=1.0]{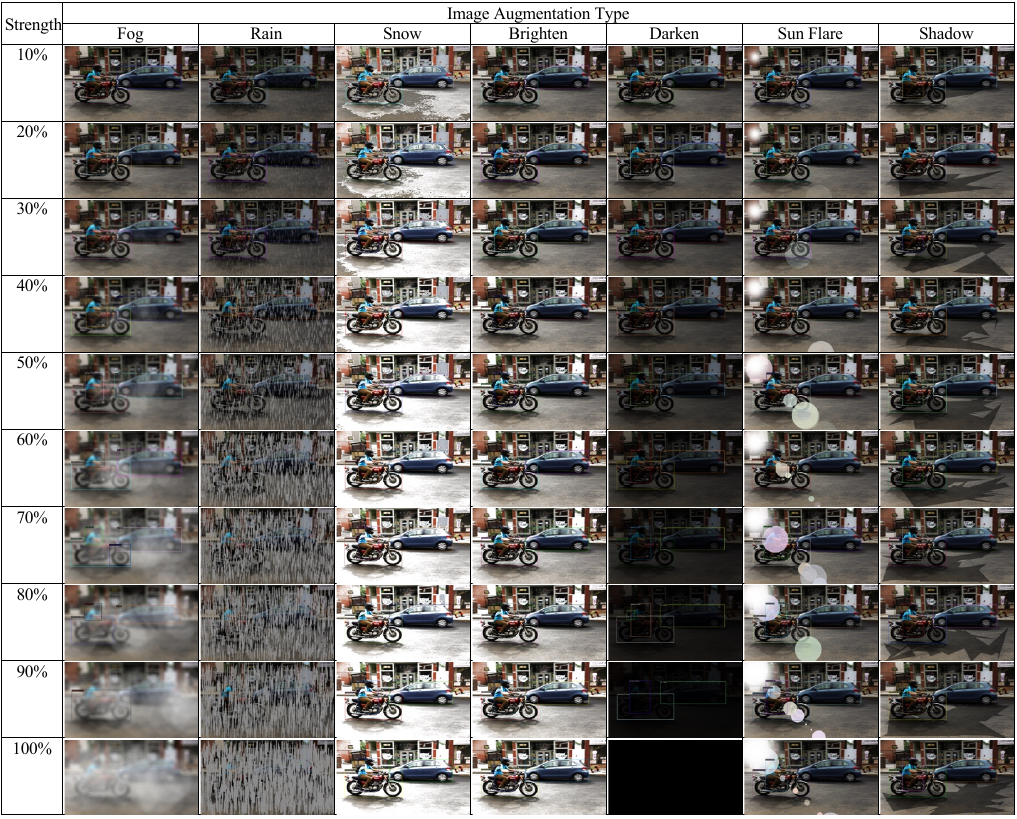}
\caption{Effects of Various Image Augmentation Operators with Interference Strengths from 10\% to 100\%}
\label{fig:ImageAugmentations}
\end{figure*}

The experiment employed 7 data augmentation operators for 7 types of adverse weather and lighting conditions, which are: Fog, Rain, Snow, Shadow, Sun Flare, Brighten, and Darken. These augmentations were implemented by the Automold road augmentation library . Fig. \ref{fig:ImageAugmentations} shows the effects of these image augmentation operations with different interference strengths from 10\% to 100\%. The original image can be found in Fig. \ref{fig:Example}.

The linear search algorithm given in Section III.C were employed to search for the first failure coefficients with the step size being 2.5\%. To avoid synthetic images being generated repeatedly in the evaluation of different models, for each image in the benchmark, a set of synthetic images was generated in advance with different severance ranging from 0 to 1 with a step size of 0.025. A total of 57,400 synthetic images were generated. 

The experiment was conducted on a desktop computer with the Windows Microsoft operating system, NVIDIA CUDA GPU (Nvidia RTX 2070 Super) and a powerful CPU (Intel i9-10850k). 

\subsubsection{Main Results}

\begin{figure}[!h]
\centering
\includegraphics[scale=1.0]{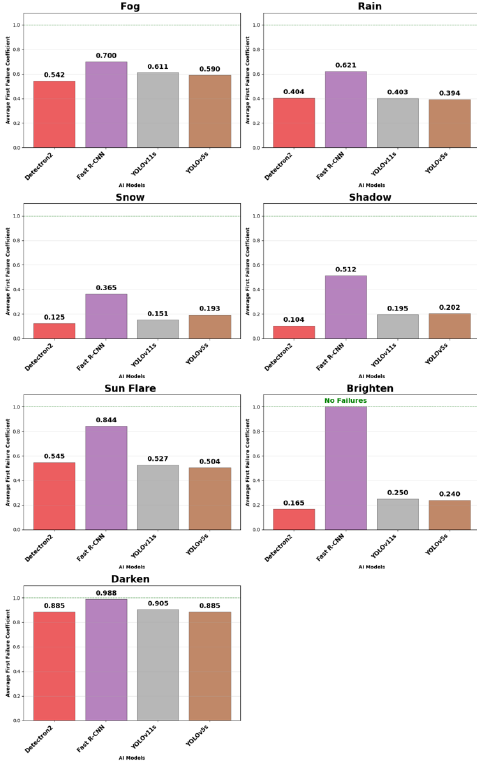}
\caption{Comparison of Average First Failure Coefficients for four object detection models in seven weather interference types}
\label{fig:Exp1ResultsChart}
\end{figure}

\begin{figure}[!h]
\centering
\includegraphics[scale=1.0]{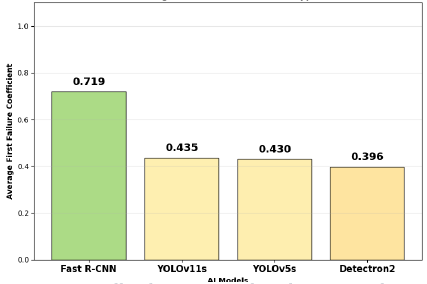}
\caption{Overall Robustness Ranking by Averaged AFFCs Across All Seven Types of Adverse Conditions}
\label{fig:Exp1OverallComparison}
\end{figure}

The results of the experiment are given in Table \ref{tab:Exp1Results} and visualised in Figure \ref{fig:Exp1ResultsChart} for the averages of the first failure coefficients (AFFC) for each ML model in every type of adverse weather conditions, while the average of the robustness scores over seven types of adverse conditions is given in Fig. \ref{fig:Exp1OverallComparison}. 

\begin{table}[!h]
\caption{Average First Failure Coefficients and The Standard Deviations for each Model-Weather Combination}
\label{tab:Exp1Results}
\includegraphics[scale=0.75]{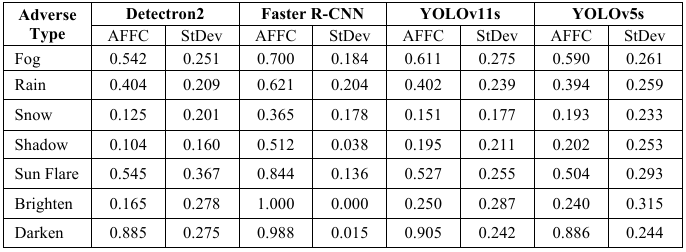}
\end{table}

The data in Fig. \ref{fig:Exp1OverallComparison} clearly shows that Faster R-CNN outperforms the other models by still being able to detect the objects in the images correctly with higher weather interference. It has demonstrated superior robustness across all weather conditions, achieving an overall AFFC of 71.9\%. This substantially exceeded the YOLO-based architectures, with YOLOv5s recording an AFFC of 43.0\% and YOLOv11s achieving 43.5\%. YOLO v11s and YOLO v5s show only a very small difference, while YOLOv11s is slightly better than YOLO v5s, as one would expect. Detectron2 is the worst among all of the models. It exhibited the lowest overall robustness among the tested models, with 39.6\%.

Analysing the data in Table \ref{tab:Exp1Results}, we can see that the newer version of YOLO models does improve upon its predecessor successfully in most types of adverse conditions, including fog, rain, sun flare, brightness and darkness. Surprisingly, YOLO v5s is more robust in snow and shadow conditions, showing that the extra training has reduced the robustness in these instances. 

As the data in Table \ref{tab:Exp1Results} shown, remarkably, the Faster R-CNN mode was completely unaffected by the brightness interference, with its ability to still detect objects correctly in every instance, even at 100\% interference strength. Similarly, with darkness, it has a 98.8\%  AFFC. The other models demonstrated strong levels of robustness, with the lowest-performing model, detectron2, achieving a higher accuracy than other interference types at 88.5\% AFFC.

The comparison between YOLOv5s and YOLOv11s revealed minimal improvement in the operational robustness despite architectural refinements. YOLOv11s outperformed its predecessor in fog (61.1\% vs 59\%), rain (40.3\% vs 39.4\%), sun flare (52.7\% vs 50.4\%), brightness (25.0\% vs 24.0\%), and darkness (90.5\% vs 88.5\%) conditions. However, YOLOv5s unexpectedly demonstrated superior robustness in snow (19.3\% vs 15.1\%) and shadow (20.2\% vs 19.5\%) conditions, resulting in an overall AFFC difference of 0.5\% between the two model versions.

Using the experiment data, one can determine the safe operating parameters of the various models. For example, Faster R-CNN’s average first failure coefficient in foggy weather conditions is 0.70. This implies that when the weather condition is above 70\% fog thickness, one can expect that Faster R-CNN will fail to recognise objects correctly. Therefore, it can be considered dangerous to rely on the object detection model. 

The data of the standard deviations provides further insight into the consistency of the pointwise robustness over the benchmark. Detectron2 exhibited particularly high variance in fog conditions with a standard deviation of 25.1\%, indicating inconsistency in the first failure points across individual images. This variation pattern was observed to varying degrees across all models and weather types, with two instances being over 30\%, suggesting that average first failure coefficients may need to be combined with the standard deviation to identify the safe operational thresholds adequately. Utilising the standard deviation can further inform manufacturers of areas of improvement when training object recognition models.

\begin{figure}[!h]
\centering
\includegraphics[scale=1.0]{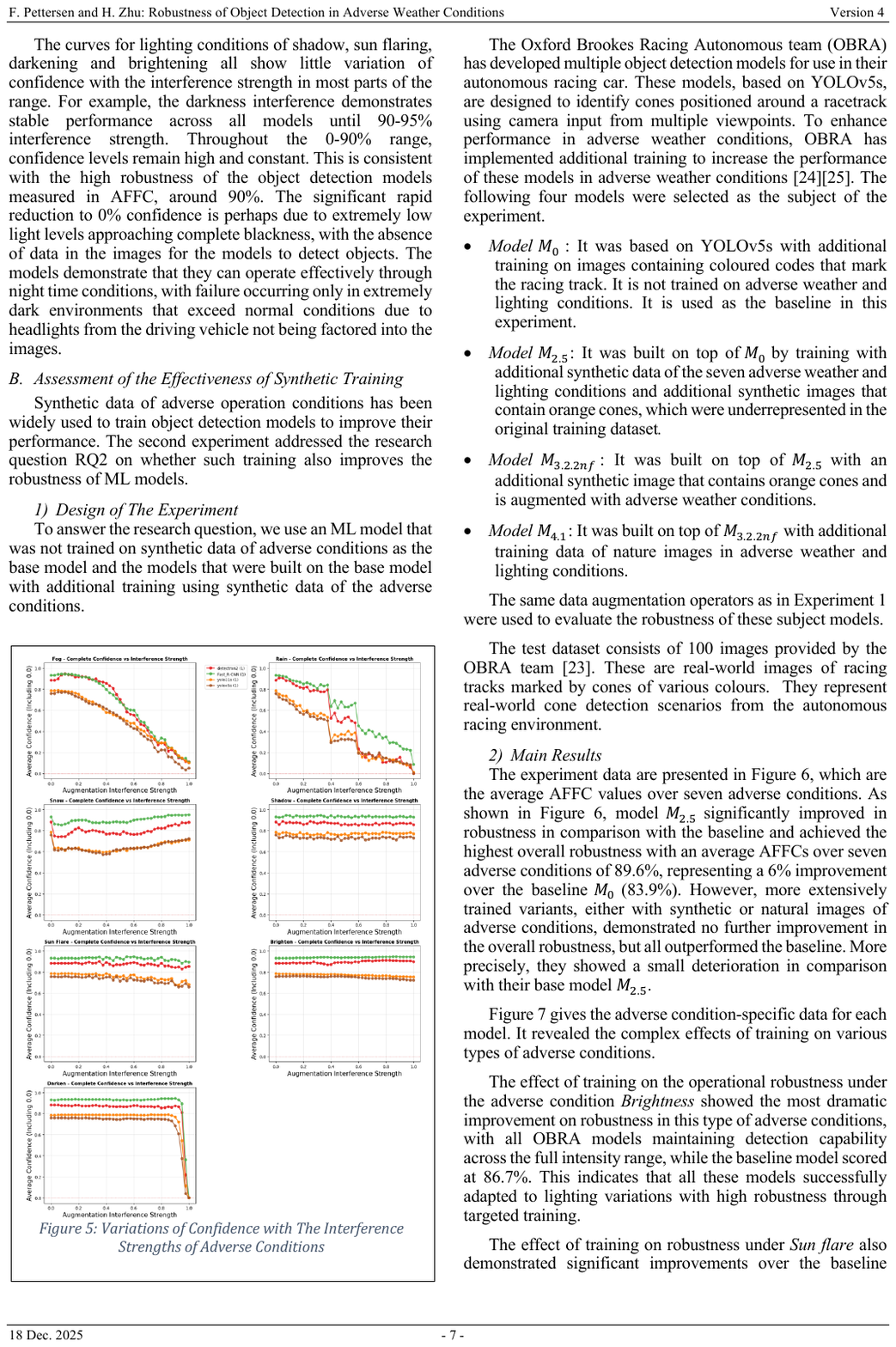}
\caption{Variations of Confidence with The Interference Strengths of Adverse Conditions}
\label{fig:ConfidenceVariations}
\end{figure}

It is worth noting that an ML model recognises objects with variable confidence. It is interesting to understand how the confidences vary with the interference strength of the adverse condition. Fig. \ref{fig:ConfidenceVariations} shows the variation of the average confidence levels of the ML model’s predictions at different interference strengths for each type of adverse condition.  

As shown in Fig. \ref{fig:ConfidenceVariations}, the fog demonstrated an ideal curve of gradual decline of confidence with the increase of interference strength, as one would expect. The confidence variation curve for rain also displays a general downward decline pattern, but contains two distinct sharp performance degradation points at approximately 40\% and 60\% interference strength, visible as sudden steep downward steps in the confidence. However, the curve for snow shows an intriguing pattern that all models experience sharp confidence drops between 0\% and 2.5\% interference strength and then climb back gradually. 

The sudden drops of confidences of the rain and snow interferences suggest that the data augmentation for rain and fog effects is not a gradual progression of interference strength, but in an abrupt way. The drops in confidence affect all models simultaneously, suggesting this is caused by the data augmentation operators rather than model-specific vulnerability.  

The curves for lighting conditions of shadow, sun flaring, darkening and brightening all show little variation of confidence with the interference strength in most parts of the range. For example, the darkness interference demonstrates stable performance across all models until 90-95\% interference strength. Throughout the 0-90\% range, confidence levels remain high and constant. This is consistent with the high robustness of the object detection models measured in AFFC, around 90\%. The significant rapid reduction to 0\% confidence is perhaps due to extremely low light levels approaching complete blackness, with the absence of data in the images for the models to detect objects. The models demonstrate that they can operate effectively through night time conditions, with failure occurring only in extremely dark environments that exceed normal conditions due to headlights from the driving vehicle not being factored into the images.

\subsection{Assessment of the Effectiveness of Synthetic Training}

Synthetic data of adverse operation conditions has been widely used to train object detection models to improve their performance. The second experiment addressed the research question RQ2 on whether such training also improves the robustness of ML models. 

\subsubsection{Design of The Experiment}

To answer the research question, we use an ML model that was not trained on synthetic data of adverse conditions as the base model and the models that were built on the base model with additional training using synthetic data of the adverse conditions. 

The Oxford Brookes Racing Autonomous team (OBRA) has developed multiple object detection models for use in their autonomous racing car. These models, based on YOLOv5s, are designed to identify cones positioned around a racetrack using camera input from multiple viewpoints. To enhance performance in adverse weather conditions, OBRA has implemented additional training to increase the performance of these models in adverse weather conditions$^{\ref{fnt:OBRA}}$. The following four models were selected as the subject of the experiment. 

\begin{itemize}
\item $Model ~M_0$: It was based on YOLOv5s with additional training on images containing coloured codes that mark the racing track. It is not trained on adverse weather and lighting conditions. It is used as the baseline in this experiment. 
\item $Model ~M_{2.5}$: It was built on top of $M_0$ by training with additional synthetic data of the seven adverse weather and lighting conditions and additional synthetic images that contain orange cones, which were underrepresented in the original training dataset. 
\item $Model ~M_{3.2.2nf}$: It was built on top of $M_{2.5}$ with an additional synthetic image that contains orange cones and is augmented with adverse weather conditions. 
\item $Model ~M_{4.1}$: It was built on top of $M_{3.2.2nf}$ with additional training data of nature images in adverse weather and lighting conditions.  
\end{itemize}
	
The same data augmentation operators as in Experiment 1 were used to evaluate the robustness of these subject models. 
The test dataset consists of 100 images provided by the OBRA team\footnote{Private Communication. \label{fnt:OBRA}}. These are real-world images of racing tracks marked by cones of various colours.  They represent real-world cone detection scenarios from the autonomous racing environment.

\subsubsection{Main Results}

\begin{figure}[!h]
\centering
\includegraphics[scale=1.0]{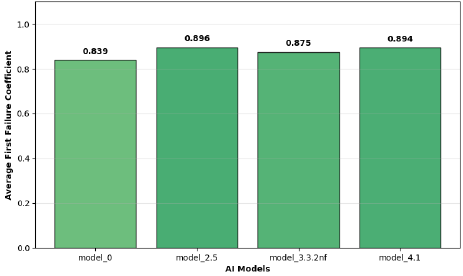}
\caption{Comparison of OBRA models on Overall Robustness Represented by the Averaged AFFC Across All Types of Adverse Conditions}
\label{fig:Exp2OverallComparison}
\end{figure}

The experiment data are presented in Fig. \ref{fig:Exp2OverallComparison}, which are the average AFFC values over seven adverse conditions. As shown in Fig. \ref{fig:Exp2OverallComparison}, model $M_{2.5}$ significantly improved in robustness in comparison with the baseline and achieved the highest overall robustness with an average AFFCs over seven adverse conditions of 89.6\%, representing a 6\% improvement over the baseline $M_0$ (83.9\%). However, more extensively trained variants, either with synthetic or natural images of adverse conditions, demonstrated no further improvement in the overall robustness, but all outperformed the baseline. More precisely, they showed a small deterioration in comparison with their base model $M_{2.5}$. 

\begin{figure}[!h]
\centering
\includegraphics[scale=1.0]{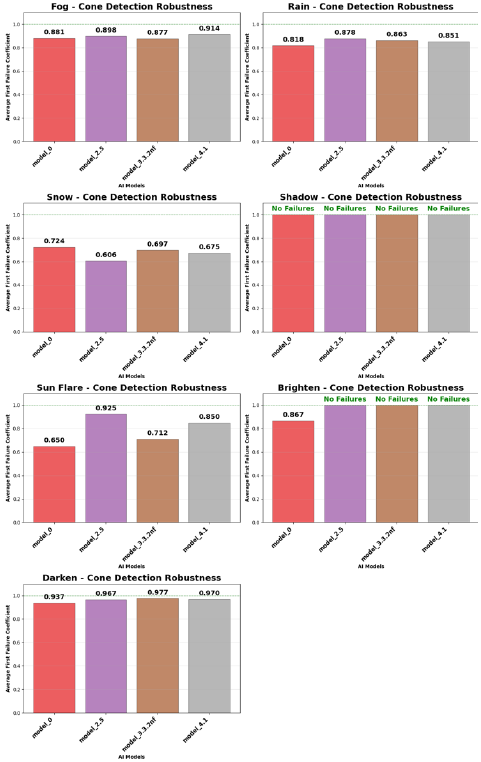}
\caption{Average First Failure Coefficient comparison across OBRA object detection models for seven weather interference conditions}
\label{fig:Exp2ResultsChart}
\end{figure}

Fig. \ref{fig:Exp2ResultsChart} gives the adverse condition-specific data for each model. It revealed the complex effects of training on various types of adverse conditions. 

The effect of training on the operational robustness under the adverse condition Brightness showed the most dramatic improvement on robustness in this type of adverse conditions, with all OBRA models maintaining detection capability across the full intensity range, while the baseline model scored at 86.7\%. This indicates that all these models successfully adapted to lighting variations with high robustness through targeted training. 

The effect of training on robustness under Sun flare also demonstrated significant improvements over the baseline model $M_0$, with model $M_{2.5}$ outperforming the rest with an AFFC of 92.5\% vs 65\% for model $M_0$.

The effect on robustness under Snow produced unexpected results, with OBRA models generally performing worse than the baseline. Model $M_0$ achieved 72.4\% AFFC, while the best-performing OBRA variant, model $M_{3.3.2nf}$, reached only 69.7\% AFFC. This degradation suggests that additional training with synthetic data may be harmful to robustness, even if the object detection performance improves. 

The data on the adverse condition Shadow revealed that all models, including the baseline, maintained the highest possible robustness. 

The data about Rain and Darkness types of adverse conditions showed OBRA models' marginal improvements (ranging from 2-5 percentage) in robustness compared to the baseline. These gains suggest that training targeting adverse conditions provided limited robustness enhancement, despite dedicated training efforts. 

The results about Fog varied inconsistently across models without a clear pattern. However, all OBRA models achieved good robustness scores, indicating that targeted training can maintain robustness at a high level. 

In summary, the experiment also demonstrates that the proposed method successfully captured subtle robustness variations between model iterations, enabling identification of specific types of adverse conditions where additional targeted training can be effective versus counterproductive. These insights provide actionable feedback for further iterative model development, highlighting which adverse conditions require refinement.

However, the findings of this experiment also challenge the assumption that the intensity of training and robustness correlates. It suggests potential diminishing returns from prolonged training. Moreover, the so-called catastrophic forgetting phenomena could also be present in robustness.

\section{Conclusion}

\subsection{Main Contributions}

In this paper, we propose a new method to test and evaluate the operational robustness of ML models in adverse operation conditions in the context of object detection models of autonomous vehicles. The experiment on four different versions of object detection models successfully demonstrates that the proposed method for quantitative measurement of the operational robustness is capable of distinguishing ML models according to their operational robustness under adverse conditions. The ability to process 57,400 weather-augmented images and to produce the AFFC metric for multiple models confirms its feasibility and efficiency. The identification of specific AFFC metric values, such as the 70\% fog intensity limit for the Faster R-CNN model, provides actionable safety information. The findings of the experiment with the four models that were trained on data of adverse conditions revealed that training on synthetic data of adverse conditions can improve the robustness of the operations in such adverse conditions. However, we also observed the phenomena of diminishing returns and catastrophic forgetting in our experiment, which suggested the limited effectiveness of training that targets adverse conditions. 

\subsection{Future Works}

There are a few limitations of the work reported in this paper that deserve further research. 

\begin{figure}
\centering
\includegraphics[scale=1.0]{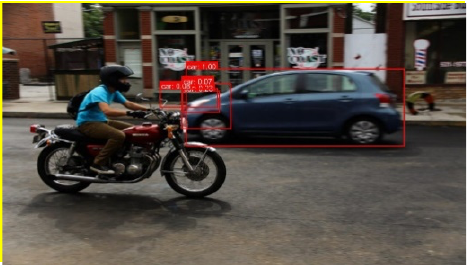}
\caption{An example that a single object is recognised as multiple objects by the Fast R-CNN Model in which the car is covered by multiple bounding boxes}
\label{fig:Example}
\end{figure}

First, the framework that we presented in the definition of the proposed method and used in experiments is a simplified model of object detection that only recognises one object in the input image. In fact, object detection ML models, such as those used in our experiments, can recognise multiple objects in each input image. In our experiments, only the best recognised object is used as the output of the ML model. This may have caused errors in the experiment results. For example, as shown in Fig. \ref{fig:Example}, a single object, i.e. the car in the image, was identified as multiple objects by the Faster R-CNN model. 

It is worth further research to extend the framework by defining the AFFC metric on ML models that detect multiple objects in the input images and further conduct experiments in such a more general setting. 

Second, weather simulation quality presents a challenge. As shown in Figure 5, the sudden confidence drops in rain at two specific thresholds, and the unusual pattern of snow indicates that the data augmentation operators for weather effects do not mirror gradual weather intensification in real environments by increasing the interference strength. The generated weather effects imperfectly replicate real-world scenarios. It is worth further research to develop data augmentation operators that better reflect the continuous change of the interference strength. 

Third, our experiments were constrained by the availability of computational resources. We have used 2.5\% of interference increase in the step size in our experiments, which enables the experiments with four models across 57,400 images to complete within 11 hours. Reducing the step size, for example, to 1\%, could produce more accurate measurements of robustness using the AFFC metric. An optimisation opportunity that could also reduce processing time is to use a more efficient search algorithm, as already pointed out in Section III.C. Thus, it is worth further study of whether ML models are monotonic in failure due to the interference strength. 

Moreover, the analysis of standard deviations revealed high variance in the interference strengths of the first failure point across individual images within the same weather condition. Detectron2's 25.1\% standard deviation in fog conditions indicates that model performance varies substantially with the features of the input images, in addition to weather intensity. Factors such as object size, distance, background, and lighting angle likely influence detection reliability in ways not captured by weather intensity alone. It is worth further research into how to take such variations into account when measuring an ML model’s operational robustness in adverse conditions. 

The current dataset size of 100 images is a small sample. As future work, the experiment test dataset should be significantly expanded, for example, to the entire COCO dataset to improve statistical validity and generalisability of results. Additionally, the scope could be broadened by testing a larger number of object recognition models to provide more options for available models that can be used in self-driving vehicles. More types of data augmentation operators for other weather and lighting conditions could be studied and used to better replicate realistic world scenarios with greater accuracy. 

The method proposed in this paper enables us to find a safety threshold in terms of the interference strength on the adverse conditions, so that above the threshold ML model of an autonomous vehicle for object detection will fail. For the threshold obtained by our method to be useful in practice, the strength of interference as the parameter of the data augmentation operator must be calibrated with the measurements of weather conditions in the real-world. For example, fog interference strength must be mapped into fog intensity as measured in terms of visibility range. Rain interference strength must be mapped into rain intensity as measured in terms of millimetres or inches of water fall per hour, and so on. 

Finally, the proposed method could be applied to a wider range of ML models beyond object detection models of autonomous vehicles. 



\bibliographystyle{IEEEtran}
\bibliography{RobustnessRefs}



\begin{IEEEbiography}[{\includegraphics
[width=1in,height=1.25in,clip,
keepaspectratio]{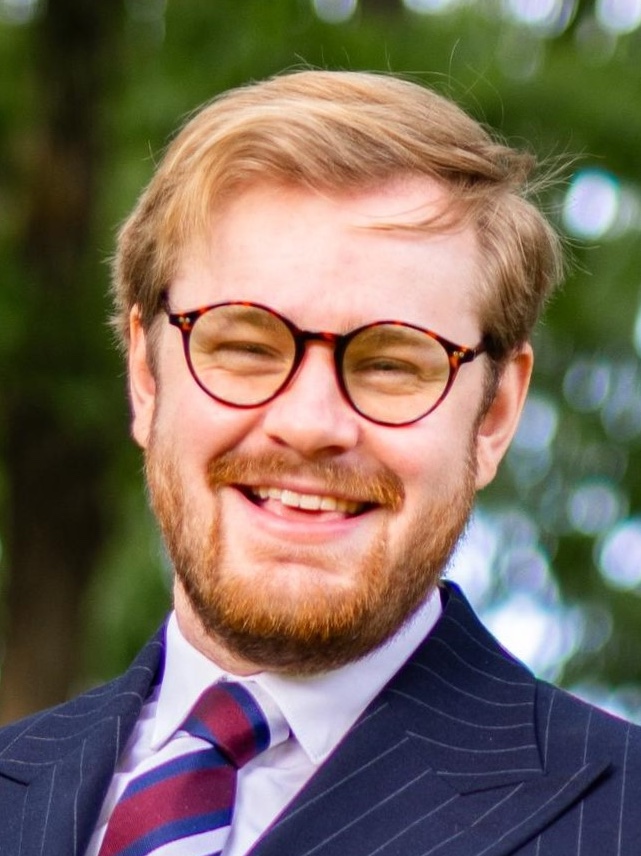}}]
{MSc Fox Pettersen} completed his postgraduate studies at Oxford Brookes University, Oxford, UK, where he studied Data Science and Artificial Intelligence. He obtained his BSc in ICT from the University of South Wales, Wales, UK, in 2024, and his MSc in Data Science and Artificial Intelligence from Oxford Brookes University in 2025. His research interests are in the area of object recognition models and their applications in automotive vehicles, including deep learning architectures for autonomous driving, real-time detection systems, and computer vision techniques for intelligent transportation systems. 
\end{IEEEbiography}

\begin{IEEEbiography}[{\includegraphics
[width=1in,height=1.25in,clip,
keepaspectratio]{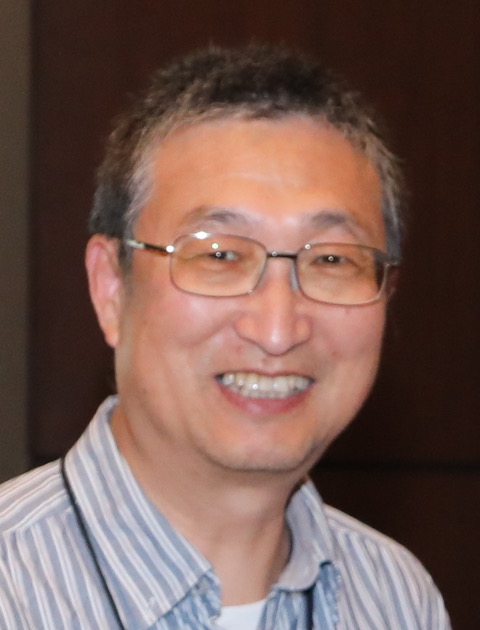}}]
{Dr. Hong Zhu} is a full professor of computer science at the Oxford Brookes University, Oxford, UK, where he chairs the Cloud Computing and Cybersecurity Research Group. He obtained his BSc, MSc and PhD degrees in Computer Science from Nanjing University, China, in 1982, 1984 and 1987, respectively. He was a faculty member of Nanjing University from 1987 to 1998 and joined Oxford Brookes University in November 1998. His research interests are in the area of software development methodologies, including software engineering of cloud-native applications, software engineering of AI and machine learning applications, formal methods, software design, software testing, programming languages, software modelling, and automated software engineering tools and environments, etc. He has published 2 books and more than 200 research papers in journals and international conferences. He is the co-founder and Conference Chair of IEEE International Conferences on AI Test and serves many international conferences in the subject area as chairs and PC members. He is a senior member of IEEE, a member of British Computer Society and ACM.
\end{IEEEbiography}
\vfill

\end{document}